\documentclass[runningheads]{llncs}
\usepackage{eccv}

\usepackage{eccvabbrv}

\usepackage{graphicx}
\usepackage{booktabs}
\usepackage{multirow}
\usepackage[table]{xcolor}
\usepackage{colortbl}

\usepackage{xcolor}
\usepackage{marvosym}

\usepackage[accsupp]{axessibility}  

\usepackage{hyperref}

\usepackage{orcidlink}

\begin{document}


\newcommand{\name}{IQA-T1\xspace}
\newcommand{\data}{Q-Tool\xspace}
\title{\name: Tool-based Visual Evidence Reasoning for Image Quality Assessment} 

\titlerunning{IQA-T1}

\author{Jinjian Wu \inst{1}*\orcidlink{0009-0009-0980-8974} \and
Jiaqi Tang\inst{2}*\orcidlink{0009-0003-1251-0825} \and
Wei Wei\inst{1}\textsuperscript{\Letter}\orcidlink{0000-0002-0655-056X} \and
Yingying Yan\inst{1}\orcidlink{0009-0007-6675-3215} \and \\
Jianmin Chen\inst{1}\orcidlink{0009-0008-4408-6930} \and
Botong Geng\inst{1} \and
Lei Zhang\inst{1}\orcidlink{0000-0002-7528-420X} \and
Qifeng Chen\inst{2}\textsuperscript{\Letter}\orcidlink{0000-0003-2199-3948}}

\authorrunning{J. Wu et al.}

\institute{
School of Computer Science, Northwestern Polytechnical University, Xi'an, China \\
\email{wujinjian@mail.nwpu.edu.cn} \ \email{weiweinwpu@nwpu.edu.cn} \and
The Hong Kong University of Science and Technology, Hong Kong SAR, China \\
\email{jtang092@connect.ust.hk} \ \email{cqf@ust.hk}
}

\maketitle

\begingroup
\renewcommand{\thefootnote}{}
\footnotetext{\textsuperscript{*} Equal contribution. \quad \textsuperscript{\Letter} Correspondence to: Wei Wei and Qifeng Chen.}
\endgroup

\begin{abstract}
Image Quality Assessment (IQA) in open-world environments remains challenging due to limited generalization and interpretability. Recent approaches based on multimodal large language models (MLLMs) introduce textual reasoning for quality prediction, yet their judgments rely heavily on semantically biased internal representations, making them insensitive to low-level perceptual degradations.
We propose \name, a tool-based visual evidence reasoning framework that augments MLLM reasoning with explicit perceptual observations. During inference, the model autonomously invokes specialized analysis tools to generate structured visual evidence, such as noise residual maps, gradient statistics, and frequency spectra, which are progressively integrated into the reasoning process.
To support this paradigm, we construct \data, a dataset containing 11k multimodal reasoning chains grounded in tool-generated evidence. Extensive experiments on seven IQA benchmarks show that \name achieves the best overall performance across datasets while producing interpretable and evidence-grounded quality assessments.
Code and dataset are available at \url{https://github.com/zibuyu-02/IQA-T1}.
\keywords{Image Quality Assessment \and Multimodal Large Language Models \and Tool-based Reasoning}
\end{abstract}

\section{Introduction}
\label{sec:intro}

Image Quality Assessment (IQA) aims to predict perceptual image quality in alignment with human subjective judgment. In image restoration~\cite{li2025mair,lin2025jarvisir,jiang2025survey,tang2026robustu1}, enhancement~\cite{an2024hfm,yan2025hvi,ma2025bilevel, Tang_2024_CVPR}, compression~\cite{he2022elic,liu2023learned,yang2023lossy}, and various downstream vision applications~\cite{tang2025intelligent}, image quality assessment not only plays an important role in algorithm performance but also directly affects the usability and robustness of vision systems in open-world environments.

\begin{figure}[t]
\centering
\includegraphics[width=1.0\linewidth]{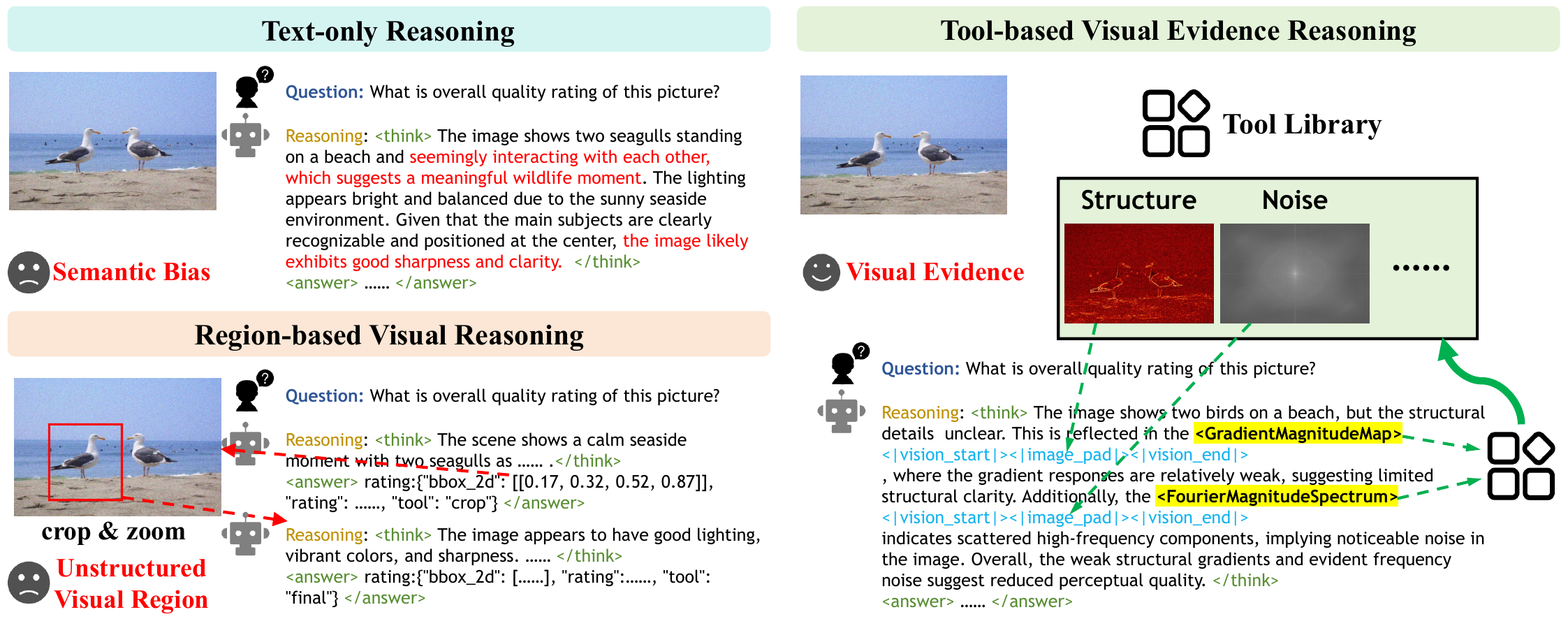}
\caption{\textbf{Motivation of our visual evidence reasoning framework.} Existing MLLM-based IQA methods typically follow two reasoning paradigms: text-only reasoning, which is prone to semantic bias, and region-based visual reasoning using cropped image patches that lack explicit perceptual interpretation.
In contrast, our framework introduces a tool-based visual evidence paradigm that generates structured visual evidence through specialized analysis tools, enabling perceptually grounded and evidence-referenced quality reasoning.}
\label{fig:Motivation}
\end{figure}

With the rapid development of multimodal large language models (MLLMs)~\cite{yang2025qwen3,touvron2023llama,li2023blip,achiam2023gpt,tang2024hawk}, researchers have begun to explore leveraging their reasoning capabilities to shift IQA from score regression toward language-driven frameworks. Although this transition promises enhanced interpretability through textual rationales and improved generalization via semantic priors, we identify a fundamental limitation shared by existing MLLM-based IQA methods: their quality assessments rely on internally represented, semantically biased visual features, rather than explicit, verifiable evidence of low-level degradation. This limitation manifests across two predominant reasoning modalities, as illustrated in Fig.~\ref{fig:Motivation}.
(1) \textbf{Text-only Reasoning.} These methods~\cite{li2025q,wu2025visualquality} generate structured quality descriptions to interpret and predict image quality. However, the reasoning process relies on the MLLM's internal representations and pre-trained semantic priors.
(2) \textbf{Region-based Visual Reasoning.} To address the lack of visual grounding in text-only reasoning, recent methods~\cite{liang2026zoom,li2026q} introduce cropped or zoomed regions of the original image during multi-step reasoning, aiming to guide the model to focus on potentially degraded areas. However, these regions are essentially unstructured pixel patches that lack explicit association with specific perceptual attributes, and thus cannot form visual evidence to support the reasoning process.

To address this fundamental gap, we introduce \name, a novel framework that instantiates a \emph{visual evidence reasoning} paradigm for IQA. The core insight is to augment the MLLM's reasoning process with explicit and structured visual evidence generated by external specialized tools. During inference, the model autonomously invokes tools from a predefined library, where each tool is designed to isolate a specific perceptual attribute, to generate intermediate visual representations such as noise residual maps, gradient magnitude distributions, and Fourier magnitude spectra. These evidence images are then incorporated into the reasoning chain as additional visual observations. This process transforms quality assessment from a purely semantic inference task into a visually grounded and evidence-driven procedure.

Realizing this paradigm requires equipping the model with two complementary capabilities: understanding \emph{how} to invoke tools and learning \emph{when} to do so strategically. We address this through a two-stage training pipeline. 
First, supervised fine-tuning (SFT) instills foundational behaviors: structured reasoning, standardized tool invocation syntax, and the ability to associate visual evidence with quality judgments. 
Second, reinforcement learning (RL) with a carefully designed reward function optimizes the tool invocation policy. The overall training reward consists of a format reward, a scoring reward, a tool usage reward, and a tool repetition penalty, which together improve scoring accuracy while suppressing redundant or improper tool calls.

Since no existing IQA dataset supports visual evidence reasoning, we construct \data, containing 11k multimodal reasoning chains grounded in tool-generated visual evidence. 
The dataset is built through a three-stage pipeline: tool library construction, visual evidence reasoning chain generation, and reasoning chain verification.

We conduct comprehensive evaluations on seven diverse IQA benchmarks. Experimental results show that \name achieves the best overall performance, with an average PLCC / SRCC of $0.795$ / $0.784$, surpassing traditional deep learning-based approaches and recent MLLM-based methods. Further ablation studies demonstrate that incorporating structured visual evidence significantly improves the stability and accuracy of quality prediction, validating the effectiveness of the visual evidence reasoning framework in enhancing both assessment reliability and interpretability.
In summary, our contributions are threefold:
\begin{itemize}
    \item We propose \name, a tool-based visual evidence reasoning framework for image quality assessment. It enables MLLMs to autonomously invoke specialized tools, dynamically constructing a structured evidence base that grounds quality assessment in explicit perceptual representations.
    \item We introduce \data, containing over 11k high-quality multimodal reasoning chains with standardized tool invocation formats and rigorous verification.
    \item We conduct extensive evaluations of \name on multiple IQA benchmarks, demonstrating its SOTA performance compared to both traditional IQA approaches and recent MLLM-based baselines.
\end{itemize}

\section{Related Work}
\label{sec:related}

\subsection{Image Quality Assessment}

Image Quality Assessment (IQA) aims to predict perceptual image quality in alignment with human subjective judgments. Traditional methods model quality degradation and output scores under either full-reference (FR-IQA)~\cite{wang2004image} or no-reference (NR-IQA)~\cite{ma2017learning,mittal2012no,mittal2012making} settings. With the rapid advancement of multimodal large language models (MLLMs)~\cite{yang2025qwen3,touvron2023llama,li2023blip,achiam2023gpt}, researchers have leveraged their cross-modal understanding to extend IQA from discriminative regression toward more generalizable, language-driven frameworks. Existing MLLM-based IQA research can be taxonomized into three evolving paradigms:  
(1) \textbf{Score-Oriented Paradigm.}
Early MLLM-based methods formulate quality prediction as classification, distribution regression, or contrastive learning to improve numerical accuracy~\cite{wang2023exploring,wu2023q,zhu2024adaptive,you2025teaching,liu2024dog}. While these approaches achieve strong correlation with human judgments, their outputs are limited to numerical values, offering no interpretability regarding why an image receives a particular score. This opacity limits their utility in applications requiring explainable decisions.
(2) \textbf{Description-Oriented Paradigm.}
To enhance interpretability, description-oriented methods~\cite{chen2024grounding,wu2024q,you2024depicting,you2025enhancing} generate structured or fine-grained natural language quality analyses. By producing textual descriptions of perceived distortions, these models provide insight into their quality judgments. However, they typically rely on supervised fine-tuning with text descriptions constructed from synthetic distortions, which fail to capture the diversity and complexity of real-world degradation patterns. Moreover, their training objectives prioritize textual coherence over numerical accuracy, often resulting in unstable or imprecise quality scores.
(3) \textbf{Textual Reasoning Paradigm.}
Recent studies have introduced explicit reasoning processes, forming a "reason-then-score" joint optimization framework. Exemplified by Q-Insight~\cite{li2025q} and VisualQuality-R1~\cite{wu2025visualquality}, these methods generate step-by-step textual rationales before predicting quality scores, often incorporating reinforcement learning to align reasoning with scoring. Despite this advancement, their quality judgments remain grounded in the MLLM's internal representations, which are inherently biased toward high-level semantics. As argued in our introduction, this semantic bias leaves these methods ill-equipped to capture low-level degradation cues, and their purely textual reasoning lacks verifiable visual anchors, rendering the generated rationales potentially disconnected from actual visual evidence.

\subsection{Thinking with Images}

The recognition that text-only reasoning lacks visual grounding has motivated a broader exploration of paradigms where images actively participate in the reasoning process. Termed "Thinking with Images"~\cite{su2025thinking}, this emerging direction transforms visual information from passive input into a dynamic cognitive workspace. Recent work can be categorized into three progressive levels~\cite{su2025thinking}: (1) tool-driven visual exploration, where models invoke external tools such as cropping and zooming to acquire fine-grained visual evidence~\cite{wang2025visuothink,zheng2025deepeyes,wang2025pixel,zhang2026cmmcot,liu2025visual}; (2) programmatic visual manipulation, where models generate executable code to perform structured edits on images~\cite{fu2025refocus,liu2025visualagentic,wang2025vrag}; and (3) intrinsic visual imagination, where models generate new visual content as intermediate reasoning steps~\cite{chen2025blip3,li2025imagine,zhao2025cot,jiang2025t2i}. These explorations mark the evolution from "thinking about images" to "thinking with images."

\noindent\textbf{Thinking with Images for IQA.}
Applying this paradigm to IQA represents a natural progression toward addressing the lack of visual evidence in text-only reasoning. Methods such as Zoom-IQA~\cite{liang2026zoom} and Q-Probe~\cite{li2026q} iteratively crop or zoom into evidence regions during inference, guiding the model to focus on locally degraded areas. While these approaches represent important steps, they remain fundamentally limited: the introduced regions are unstructured pixel patches that still rely on the model's semantic interpretation. They do not generate explicit, structured visual evidence that directly highlights specific perceptual attributes such as noise characteristics, frequency anomalies, or structural coherence. Consequently, they fail to bridge the representational gap between semantic reasoning and low-level degradation, which motivates our proposed \name framework based on tool-generated visual evidence.

\section{Dataset}
\label{sec:dataset}

\begin{figure}[t]
\centering
\includegraphics[width=1.0\linewidth]{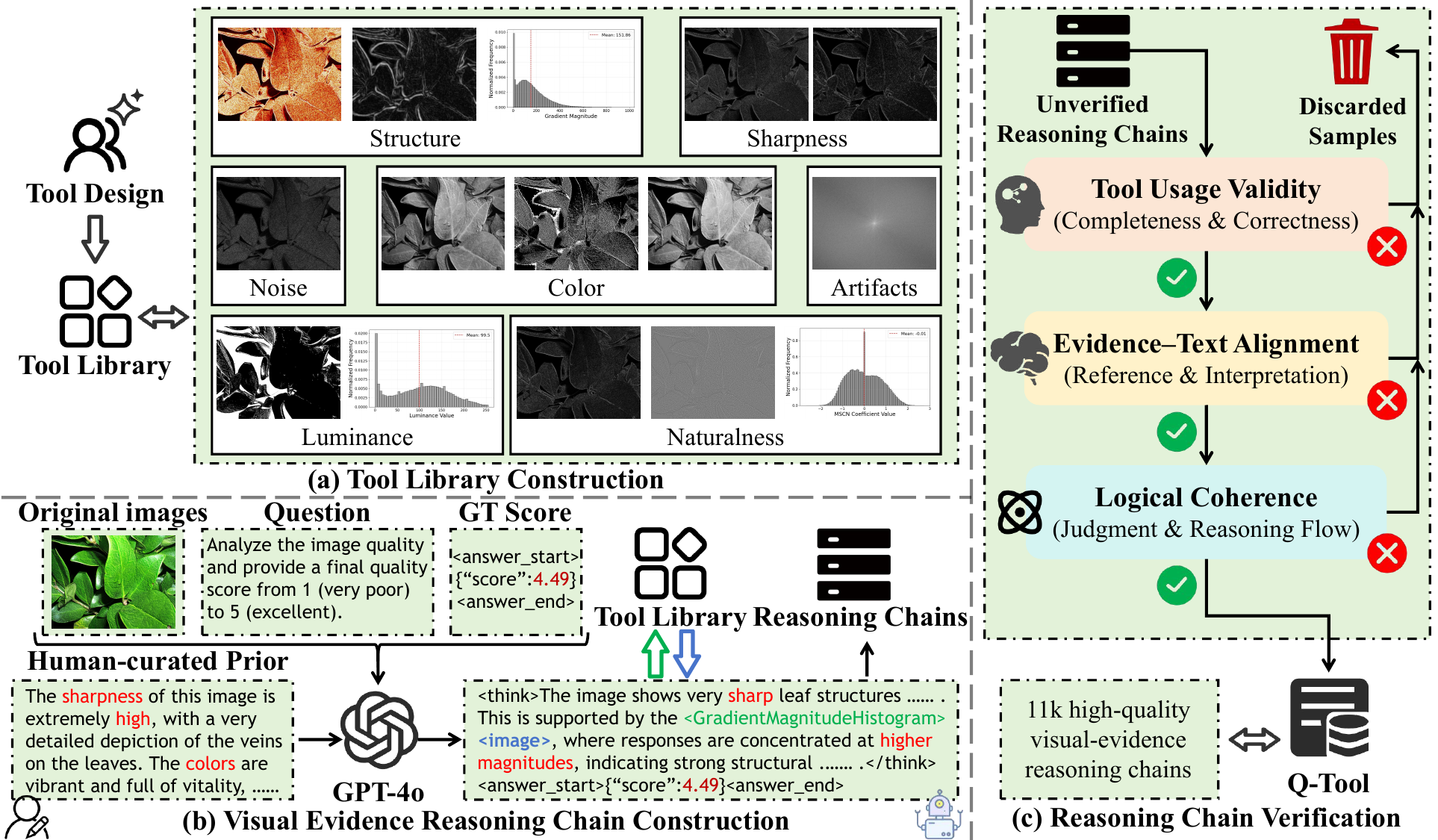}
\caption{Overview of the \data dataset construction pipeline, consisting of three key stages: 
(a) Perceptual tool library construction for visual evidence generation; 
(b) Visual evidence-grounded reasoning chain synthesis guided by human-curated priors; 
(c) Multi-level verification of reasoning chains.}
\label{fig:dataset_pipeline}
\end{figure}

Existing MLLM-based IQA methods suffer from a fundamental limitation: their quality assessments rely on semantically biased internal representations rather than explicit, verifiable evidence of low-level degradation. Addressing this gap requires a dataset that explicitly models the relationship between perceptual degradation and structured visual evidence within a reasoning framework. However, no such dataset currently exists. To enable the training of visual evidence reasoning models, we introduce \data—the first dataset specifically designed for tool-based visual evidence reasoning in IQA. It comprises 11k high-quality multimodal reasoning chains, each integrating explicit tool-generated visual evidence with human-aligned textual rationales.

The construction of \data is guided by three requirements: (1) the dataset must include explicit visual evidence that captures low-level perceptual attributes; (2) the reasoning chains must systematically integrate this evidence with textual analysis; and (3) the overall structure must support supervised fine-tuning of MLLMs to learn evidence-grounded reasoning behaviors. To meet these requirements, we design a systematic three-stage construction pipeline, illustrated in Fig.~\ref{fig:dataset_pipeline}. The following sections describe each stage in detail.

\subsection{Tool Library Construction}
\label{sec:tool_design}

The foundation of visual evidence reasoning is the ability to generate explicit representations of low-level perceptual attributes. However, as argued in Sec.~\ref{sec:intro}, MLLM internal representations are inherently biased toward high-level semantics and fail to capture degradation cues such as noise, blur, or frequency anomalies. To overcome this limitation, we construct a library of specialized perceptual tools, each designed to isolate and visualize a specific quality-relevant attribute.

\noindent\textbf{Design rationale.} We identify seven perceptual attributes critical for IQA, including structure, sharpness, noise, artifacts, luminance, color, and naturalness. Based on these attributes, we design 15 corresponding types of visual evidence. Detailed descriptions are provided in the supplementary material. Each tool \(T_k\) maps an input image \(I\) to a structured visual representation \(e_k = \psi(T_k(I))\) that highlights degradation patterns invisible to standard MLLM representations. For example, the \textit{NoiseResidualMap} isolates sensor or compression noise by subtracting a denoised version from the original; the \textit{FourierMagnitudeSpectrum} reveals periodic artifacts through frequency-domain analysis; and \textit{GradientOrientationCoherenceMap} exposes structural inconsistencies caused by blur or over-smoothing.

\noindent\textbf{Implementation considerations.} To ensure stable training and inference, all tools share three properties. First, they adopt unified invocation interfaces, allowing the MLLM to call any tool using a consistent syntax. Second, outputs are moderately downsampled to balance spatial structure preservation with visual token efficiency. Third, all tools are deterministic, guaranteeing identical outputs across data construction, supervised fine-tuning, and reinforcement learning stages—a critical requirement for consistent policy learning.

This tool library provides the perceptual foundation for subsequent reasoning chain construction, enabling the explicit representation of degradation cues that semantic MLLM representations inherently lack.

\subsection{Visual Evidence Reasoning Chain Construction}
\label{sec:data_synthesis}

With the tool library established, we next construct multimodal reasoning chains that systematically integrate visual evidence into quality assessment. A reasoning chain consists of (i) explicit tool invocations at appropriate analytical stages, (ii) interpretation of the generated visual evidence, and (iii) a final quality judgment logically derived from the accumulated evidence.

\noindent\textbf{Challenge of naive generation.} Directly prompting large language models to generate such chains autonomously leads to semantic hallucinations, logical discontinuities, and improper tool usage. The resulting chains lack the structured, verifiable relationship between evidence and judgment required for effective supervision.

\noindent\textbf{Human-curated reasoning templates.} To address this, we first develop human-curated reasoning templates that encode expert knowledge about perceptual analysis. These templates define: the logical decomposition of quality assessment into stages (e.g., "examine structure," "analyze noise," "assess color fidelity"); the mapping from perceptual attributes to appropriate tools; and the causal relationships between observed evidence and final quality judgment. These templates serve as semantic constraints, ensuring that generated chains follow human-aligned perceptual logic.

\noindent\textbf{Constrained generation with GPT-4o.} Under these constraints, we employ GPT-4o to generate complete multimodal reasoning chains. For each image, the model receives: the image itself, the corresponding reasoning template, and access to the tool library. At each designated stage, the model inserts explicit tool invocations, and the resulting visual evidence is integrated into subsequent analysis. This process transforms purely textual reasoning into an evidence-grounded procedure where each inference step relies on observable visual signals.

By combining human perceptual priors with controlled model generation, we produce reasoning chains that are both logically structured and grounded in verifiable evidence.

\subsection{Reasoning Chain Verification}
\label{sec:data_quality_control}

The reliability of \data as a training resource depends on the quality of its constituent reasoning chains. We therefore conduct systematic multi-level verification to ensure consistent alignment among tool usage, visual evidence, and reasoning logic.

\noindent\textbf{Verification protocol.} Each generated chain is examined along three dimensions. (1) \textbf{Tool appropriateness}: Are tool invocations aligned with the intended reasoning stage and analytical purpose? We verify that tools are neither redundant nor improperly used. (2) \textbf{Evidence grounding}: Is the generated visual evidence correctly referenced, interpreted, and integrated into the textual reasoning? We ensure that each evidence image serves a genuine analytical function rather than appearing as a decorative element. (3) \textbf{Logical consistency}: Is the final quality judgment sufficiently supported by the preceding evidence accumulation? We reject chains where conclusions lack adequate evidentiary support or exhibit reasoning discontinuities.

\noindent\textbf{Quality refinement.} Chains failing any verification criterion are removed. A subset of borderline cases undergoes manual refinement to correct minor inconsistencies while preserving the overall structure. This rigorous verification process yields a final dataset of 11k high-quality reasoning chains, each exhibiting: (i) appropriate tool utilization, (ii) coherent integration of visual evidence, and (iii) logically sound quality judgments.

The resulting \data dataset provides the necessary supervision for learning evidence-grounded reasoning behaviors in the subsequent two-stage training pipeline (Sec.~\ref{sec:sft} and Sec.~\ref{sec:rl}), serving as the foundation for the visual evidence reasoning paradigm introduced in this work.

\section{Methodology}
\label{sec:method}

\begin{figure}[t]
\centering
\includegraphics[width=1\linewidth]{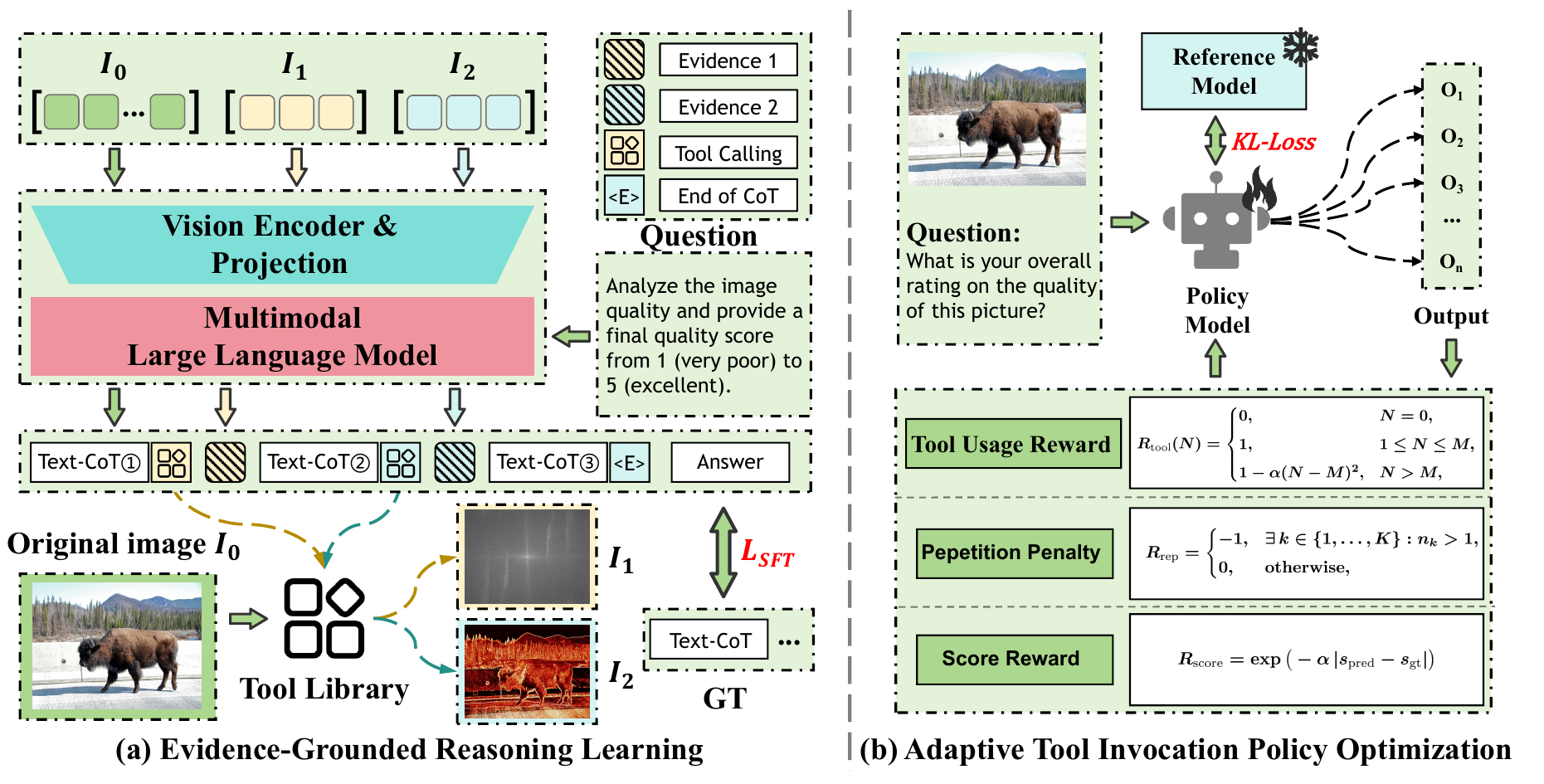}
\caption{Overview of the proposed \name framework. The model operates in two stages: (1) Supervised fine-tuning (SFT) on the \data dataset to learn evidence-grounded reasoning and tool invocation syntax; (2) Reinforcement learning (RL) with a tailored reward function to optimize the adaptive tool invocation policy. During inference, the model autonomously invokes tools to acquire visual evidence $\mathbf{E}_r$, which is integrated into the reasoning chain to ground quality assessment.}
\label{fig:framework}
\end{figure}

In this section, we first formalize the problem of MLLM-based IQA and highlight the semantic bias inherent in standard visual representations (Sec.~\ref{sec:problem}). We then introduce the proposed visual evidence reasoning framework (Sec.~\ref{sec:ver}) and describe the two-stage training pipeline comprising supervised fine-tuning (Sec.~\ref{sec:sft}) and reinforcement learning (Sec.~\ref{sec:rl}) to equip the model with both foundational reasoning capabilities and an adaptive tool invocation policy.

\subsection{Problem Formulation}
\label{sec:problem}

\noindent\textbf{MLLM-based IQA as Sequence Generation.}
We formalize image quality assessment using a multimodal large language model (MLLM) as a conditional sequence generation problem. Let $I \in \mathbb{R}^{H \times W \times 3}$ denote the input image and $P$ the textual prompt (e.g., ``Assess the quality of this image''). The target output is a sequence $Y = \{y_1, y_2, \dots, y_T\}$ that includes both a reasoning chain and a final quality score $s \in \mathbb{R}$ (typically normalized to $[1,5]$).
A standard MLLM comprises a visual encoder $\mathcal{E}_\phi$, a modality projector $\mathcal{M}$, and a large language model (LLM) $\mathcal{G}_\theta$. The image is first encoded and projected into a sequence of visual tokens aligned with the textual embedding space: $\mathbf{z}_v = \mathcal{M}\big(\mathcal{E}_\phi(I)\big) \in \mathbb{R}^{L \times d},$ where $L$ is the number of visual tokens and $d$ the embedding dimension. The inference process then models the posterior distribution of the output sequence conditioned on the multimodal inputs:
\begin{equation}
    p(Y \mid I, P) = \prod_{t=1}^{T} p_\theta(y_t \mid \mathbf{z}_v, \mathbf{h}_p, y_{<t}),
    \label{eq:vlm_inference}
\end{equation}
with $\mathbf{h}_p$ denoting the prompt embeddings and $y_{<t}$ the previously generated tokens.

\noindent\textbf{Semantic Bias in Visual Representations.}
A growing body of work~\cite{chen2026mitigating,li2025investigate} has shown that the visual representation $\mathbf{z}_v$ produced by MLLMs is heavily biased toward high-level semantic content, while offering limited sensitivity to low-level perceptual degradations critical for IQA, such as noise, blur, or compression artifacts. Formally, consider an image $I$ and its perceptually degraded counterpart $I_d$ (e.g., with added noise). Although human observers would assign significantly different quality scores $s(I)$ and $s(I_d)$, the corresponding visual representations often remain nearly identical:
\begin{equation}
\|\mathbf{z}_v(I) - \mathbf{z}_v(I_d)\|_2 \approx 0, \qquad 
|s(I) - s(I_d)| \gg \epsilon,
\label{eq:semantic_bias}
\end{equation}
where $\epsilon$ is a small tolerance. This observation reveals that $\mathbf{z}_v$ alone is insufficient to capture the degradation cues necessary for accurate quality assessment, motivating the introduction of complementary perceptual evidence.

\subsection{Visual Evidence Reasoning}
\label{sec:ver}

To compensate for the information loss in $\mathbf{z}_v$, we propose a framework that augments the MLLM's reasoning process with explicit, structured visual evidence generated by external tools. Let $\mathcal{T} = \{T_k\}_{k=1}^K$ denote a library of specialized perceptual analysis tools, each mapping an image to a degradation-oriented feature space: $T_k : \mathbb{R}^{H \times W \times 3} \to \mathcal{D}_k,$ where $\mathcal{D}_k$ is the output domain (e.g., a residual map, histogram, or spectrum). The set of all possible visual evidence for image $I$ is defined as
\begin{equation}
\mathbf{E} = \{e_k \mid e_k = \psi(T_k(I)),\; k=1,\dots,K\},
\end{equation}
with $\psi(\cdot)$ projecting tool outputs into the LLM's context space (e.g., downsampling and tokenization).

During inference, the model does not access the entire set $\mathbf{E}$ upfront. Instead, it dynamically constructs a subset $\mathbf{E}_r \subseteq \mathbf{E}$ by invoking relevant tools at appropriate reasoning steps. The generation process thus conditions on both the original visual tokens and the progressively acquired evidence:
\begin{equation}
\hat{Y} = \operatorname*{arg\,max}_{Y} \sum_{t=1}^{T} \log p_\theta(y_t \mid \mathbf{z}_v, \mathbf{h}_p, \mathbf{E}_r, y_{<t}).
\label{eq:evidence_inference}
\end{equation}

The evidence $\mathbf{E}_r$ serves as a dynamically acquired perceptual complement to $\mathbf{z}_v$, enabling the model to ground each reasoning step in verifiable, low-level visual cues. This formulation transforms IQA from a purely semantic inference task into an evidence-driven reasoning process.

\subsection{Two-Stage Training Pipeline}

Realizing the visual evidence reasoning paradigm requires equipping the MLLM with two complementary capabilities: (i) understanding \emph{how} to invoke tools and interpret the resulting evidence, and (ii) learning \emph{when} to invoke tools strategically to maximize assessment accuracy while avoiding redundancy. We address these via a two-stage pipeline: supervised fine-tuning (SFT) on the \data dataset establishes foundational reasoning behaviors, followed by reinforcement learning (RL) that optimizes the adaptive tool invocation policy.

\noindent\textbf{Stage I: Evidence-Grounded Reasoning Learning}
\label{sec:sft}

In this stage, we train the model on the \data dataset to learn structured, evidence-grounded reasoning chains. Each training sample consists of the original image $I$, a set of tool-generated visual evidence $V$ (corresponding to the ground-truth invocations), and a textual reasoning chain $T$ that integrates the evidence and culminates in a quality score $s_{\text{gt}}$.

Crucially, the model is not required to generate the visual evidence itself, as the evidence is deterministically produced by the tools and provided as input. During SFT, we apply loss masking to the tokens representing the evidence images, as they are not prediction targets. The objective is to maximize the likelihood of the reasoning tokens conditioned on the complete context:
\begin{equation}
\mathcal{L}_{\text{SFT}} = - \sum_{t=1}^{L} \log P_\theta(y_t \mid y_{<t}, I, V, T),
\label{eq:sft_loss}
\end{equation}
where $y_t$ denotes the $t$-th ground-truth token of the reasoning sequence (including the final score) and $L$ is its length.
Optimizing Eq.~\eqref{eq:sft_loss} instills three essential behaviors: (i) the ability to follow structured reasoning templates, (ii) correct syntax for tool invocation, and (iii) the capacity to associate visual evidence with quality judgments. This stage initializes a policy $\pi_{\text{SFT}}$ that serves as the foundation for subsequent RL fine-tuning.

\noindent\textbf{Stage II: Adaptive Tool Invocation Policy Optimization}
\label{sec:rl}

While SFT establishes correct reasoning patterns, it does not explicitly optimize the \emph{strategy} of when to invoke which tool. To learn an adaptive invocation policy, we employ reinforcement learning with a reward function designed to balance prediction accuracy, evidence efficiency, and reasoning stability.

\noindent\textbf{Reinforcement Learning Setup.}
We adopt Group Relative Policy Optimization (GRPO)~\cite{guo2025deepseek}, a variant of PPO that estimates advantages from within-group comparisons, eliminating the need for a separate value network~\cite{tang2026lpo}. Given a prompt $x$, the current policy $\pi_\theta$ samples a group of $n$ reasoning chains $\{y_1,\dots,y_n\}$. The objective is:
\begin{equation}
\mathcal{L}_{\text{GRPO}}(\theta) = - \mathbb{E}_{x \sim \mathcal{D},\, y_i \sim \pi_\theta(\cdot|x)}
\Big[ \hat{A}_i \log \pi_\theta(y_i|x) \Big]
+ \beta \, D_{\mathrm{KL}}\!\left(\pi_\theta \,\|\, \pi_{\text{ref}}\right),
\label{eq:grpo}
\end{equation}
where $\hat{A}_i = (R_i - \mu_R)/\sigma_R$ is the normalized advantage within the group, $\pi_{\text{ref}}$ is the reference policy from Stage I, and $\beta$ controls the KL penalty to prevent catastrophic forgetting.

\noindent\textbf{Reward Design.}
The reward function $R(x,y)$ is a weighted combination of four components:

(1) \textbf{Format reward} $R_{\text{fmt}}$: enforces that the output adheres to the expected structured format (e.g., contains tool invocation tags and a final score). It is $1$ if the format is correct, $0$ otherwise.

(2) \textbf{Scoring reward} $R_{\text{score}}$: measures accuracy of the predicted quality score $s_{\text{pred}}$ relative to the ground truth $s_{\text{gt}}$: 
\begin{equation}
R_{\text{score}} = \exp\big(-\alpha \, |s_{\text{pred}} - s_{\text{gt}}|\big),
\end{equation}
with $\alpha>0$ controlling the sensitivity.

(3) \textbf{Tool usage reward} $R_{\text{tool}}$: encourages efficient evidence acquisition by rewarding an appropriate number of tool invocations. Let $N = |\mathbf{E}_r|$ be the number of distinct evidence items collected. Then
    \begin{equation}
    R_{\text{tool}}(N)=
    \begin{cases}
    0, & N = 0,\\[4pt]
    1, & 1 \le N \le M,\\[4pt]
    1 - \gamma (N-M)^2, & N > M,
    \end{cases}
    \label{eq:tool_reward}
    \end{equation}
    where $M$ is a predefined maximum (set to $4$ in our experiments) and $\gamma$ controls the penalty for excessive invocations.This reward is intentionally designed as a weak constraint rather than a strict tool-level supervision signal, since real-world IQA images often involve mixed degradations and may admit multiple valid tool combinations. Therefore, $R_{\text{tool}}$ encourages the model to acquire sufficient visual evidence while preserving flexibility in adaptive tool selection.
    
(4) \textbf{Repetition penalty} $R_{\text{rep}}$: prevents degenerate behavior where the model repeatedly invokes the same tool without gaining new information:
    \begin{equation}
R_{\text{rep}} =
\begin{cases}
-1, & \exists\, k \in \{1,\dots,K\} : n_k > 1, \\[1mm]
0, & \text{otherwise},
\end{cases}
\label{eq:rep_penalty}
\end{equation}
    where $n_k$ is the number of invocations of tool $T_k$.

The overall reward is
\begin{equation}
R = \lambda_{\text{fmt}} R_{\text{fmt}} + \lambda_{\text{score}} R_{\text{score}} + \lambda_{\text{tool}} R_{\text{tool}} + \lambda_{\text{rep}} R_{\text{rep}},
\label{eq:total_reward}
\end{equation}
with weights $\lambda$ balancing the contributions (set empirically as $\lambda_{\text{fmt}}=1$, $\lambda_{\text{score}}=5$, $\lambda_{\text{tool}}=1$, $\lambda_{\text{rep}}=2$).
Through this reward formulation, the RL stage refines the tool invocation policy to maximize scoring accuracy while maintaining efficient and non-redundant use of perceptual evidence, ultimately yielding a model whose reasoning is both accurate and interpretable.

\section{Experiments}
\label{sec:experiments}

\begin{table*}[t]
\centering
\Huge
\caption{PLCC/SRCC comparison on score regression tasks between our method and other competitive IQA methods. All methods except handcrafted ones are trained on the KonIQ dataset. AVG. denotes the average PLCC/SRCC across all evaluation datasets. The best result is bolded and the second best is underlined.}
\label{tab:plcc_srcc}

\renewcommand{\arraystretch}{1.15} 

\resizebox{\textwidth}{!}{
\begin{tabular}{l c|c|c | c|c|c | c | c}
\toprule
\multirow{2}{*}{\textbf{Methods}}
& \multicolumn{3}{c}{\textbf{Wild Images}}
& \multicolumn{3}{c}{\textbf{Synthetic Distortion}}
& \multicolumn{1}{c}{\textbf{AI-Generated}}
& \multirow{2}{*}{\textbf{AVG.}} \\
\cmidrule(lr){2-4}
\cmidrule(lr){5-7}
\cmidrule(lr){8-8}
& \multicolumn{1}{c}{KonIQ~\cite{hosu2020koniq}}
& \multicolumn{1}{c}{SPAQ~\cite{fang2020perceptual}}
& \multicolumn{1}{c}{LiveW~\cite{ghadiyaram2015live}}
& \multicolumn{1}{c}{KADID~\cite{lin2019kadid}}
& \multicolumn{1}{c}{PIPAL~\cite{jinjin2020pipal}}
& \multicolumn{1}{c}{CSIQ~\cite{larson2010most}}
& \multicolumn{1}{c}{AGIQA~\cite{li2023agiqa}}
& \\
\midrule

\multicolumn{9}{c}{\textbf{Handcrafted Methods}} \\
\midrule
NIQE~\cite{mittal2012making}
& 0.533 / 0.530 & 0.679 / 0.664 & 0.493 / 0.449
& 0.468 / 0.405 & 0.195 / 0.161 & 0.718 / 0.628
& 0.560 / 0.533
& 0.521 / 0.481 \\
BRISQUE~\cite{mittal2012no}
& 0.225 / 0.226 & 0.490 / 0.406 & 0.361 / 0.313
& 0.429 / 0.356 & 0.267 / 0.232 & 0.740 / 0.556
& 0.541 / 0.497
& 0.436 / 0.369 \\

\midrule
\multicolumn{9}{c}{\textbf{Non-MLLM Methods}} \\
\midrule
NIMA~\cite{talebi2018nima}
& 0.896 / 0.859 & 0.838 / 0.856 & 0.814 / 0.771
& 0.532 / 0.535 & 0.390 / 0.399 & 0.695 / 0.649
& 0.715 / 0.654
& 0.697 / 0.675 \\
HyperIQA~\cite{su2020blindly}
& 0.917 / 0.906 & 0.791 / 0.788 & 0.772 / 0.749
& 0.506 / 0.468 & 0.410 / 0.403 & 0.752 / 0.717
& 0.702 / 0.640
& 0.693 / 0.667 \\
DBCNN~\cite{zhang2018blind}
& 0.884 / 0.875 & 0.812 / 0.806 & 0.773 / 0.755
& 0.497 / 0.484 & 0.384 / 0.381 & 0.586 / 0.572
& 0.730 / 0.641
& 0.667 / 0.645 \\
MUSIQ~\cite{ke2021musiq}
& 0.924 / 0.929 & 0.868 / 0.863 & 0.789 / 0.830
& 0.575 / 0.556 & 0.431 / 0.431 & 0.771 / 0.710
& 0.722 / 0.630
& 0.726 / 0.707 \\
ManIQA~\cite{yang2022maniqa}
& 0.849 / 0.834 & 0.768 / 0.758 & 0.849 / 0.832
& 0.499 / 0.465 & 0.457 / 0.452 & 0.623 / 0.627
& 0.723 / 0.636
& 0.681 / 0.658 \\

\midrule
\multicolumn{9}{c}{\textbf{MLLM-based Methods (w/o reasoning)}} \\
\midrule
CLIP-IQA+~\cite{wang2023exploring}
& 0.909 / 0.895 & 0.866 / 0.864 & 0.832 / 0.805
& 0.653 / 0.654 & 0.427 / 0.419 & 0.772 / 0.719
& 0.736 / 0.685
& 0.742 / 0.720 \\
C2Score~\cite{zhu2024adaptive}
& 0.923 / 0.910 & 0.867 / 0.860 & 0.786 / 0.772
& 0.500 / 0.453 & 0.354 / 0.342 & 0.735 / 0.705
& 0.777 / 0.671
& 0.706 / 0.673 \\
Q-Align~\cite{wu2023q}
& 0.941 / \underline{0.940} & 0.886 / 0.887 & 0.853 / 0.860
& 0.674 / 0.684 & 0.403 / 0.419 & 0.671 / 0.737
& 0.772 / 0.735
& 0.743 / 0.752 \\
DeQA~\cite{you2025teaching}
& \textbf{0.953} / \textbf{0.941} & 0.895 / 0.896 & \textbf{0.892} / \textbf{0.879}
& 0.694 / 0.687 & \underline{0.472} / \underline{0.478} & 0.787 / 0.744
& 0.809 / 0.729
& 0.786 / 0.765 \\

\midrule
\multicolumn{9}{c}{\textbf{MLLM-based Methods (w/ reasoning)}} \\
\midrule
Q-Insight~\cite{li2025q}
& 0.918 / 0.895 & \underline{0.903} / \underline{0.903} & 0.870 / 0.839
& \underline{0.702} / 0.702 & 0.458 / 0.435 & 0.685 / 0.640
& \underline{0.816} / \textbf{0.766}
& 0.765 / 0.740 \\
VisualQuality-R1~\cite{wu2025visualquality}
& 0.910 / 0.896 & 0.889 / 0.892 & 0.856 / 0.827
& \textbf{0.703} / \textbf{0.712} & 0.451 / 0.441 & 0.768 / 0.707
& \textbf{0.817} / 0.760
& 0.771 / 0.748 \\
Zoom-IQA~\cite{liang2026zoom}
& 0.938 / 0.922 & 0.902 / 0.900 & 0.887 / \underline{0.870}
& 0.701 / 0.700 & 0.468 / 0.465 & \underline{0.797} / \underline{0.754}
& \underline{0.816} / \underline{0.765}
& \underline{0.787} / \underline{0.768} \\
\name
& \underline{0.942} / 0.925 & \textbf{0.905} / \textbf{0.908} & \underline{0.888} / 0.860
& 0.686 / \underline{0.708} & \textbf{0.480} / \textbf{0.490} & \textbf{0.850} / \textbf{0.840}
& \underline{0.816} / 0.754
& \textbf{0.795} / \textbf{0.784} \\

\bottomrule
\end{tabular}
}
\end{table*}

\subsection{Experimental Setup}
\label{sec:setup}

\noindent\textbf{Datasets and Metrics.} 
To evaluate the robustness of \name under diverse degradations, we follow prior works by training the model on KonIQ~\cite{hosu2020koniq} and evaluating its cross-distribution generalization on six IQA benchmarks. KonIQ~\cite{hosu2020koniq}, SPAQ~\cite{fang2020perceptual}, and LIVE-Wild~\cite{ghadiyaram2015live} represent real-world distortions; KADID~\cite{lin2019kadid} and CSIQ~\cite{larson2010most} contain synthetic distortions; and PIPAL~\cite{jinjin2020pipal} focuses on algorithm-induced degradations. 
We further include AGIQA~\cite{li2023agiqa} to evaluate AI-generated content. 
Following common IQA practice, performance is measured using PLCC and SRCC.

\noindent\textbf{Implementation Details.}
We build our framework upon Qwen3-VL-4B and adopt a two-stage training pipeline. In the first stage, we perform SFT on our \data datasets using cross-entropy loss. Training is conducted with a batch size of 1, 16 gradient accumulation steps, a learning rate of $1\times10^{-5}$, and a warm-up ratio of 0.05 for two epochs. In the second stage, we apply GRPO to further align the model with perceptual quality preferences. The GRPO training initializes from the SFT checkpoint and uses a global batch size of 12 with a micro-batch size of 1 per GPU. The learning rate is set to $1\times10^{-6}$, and each prompt generates $N=8$ rollout samples. Training is performed on six NVIDIA RTX 4090 GPUs, and the KL-penalty coefficient is fixed at $1\times10^{-2}$.

\noindent\textbf{Baseline Methods.}
We evaluate \name on the image quality score regression task against four categories of baselines. 
(1) Traditional handcrafted IQA metrics, including NIQE~\cite{mittal2012making} and BRISQUE~\cite{mittal2012no}. 
(2) Deep learning-based IQA models such as NIMA~\cite{talebi2018nima}, HyperIQA~\cite{su2020blindly}, DBCNN~\cite{zhang2018blind}, MUSIQ~\cite{ke2021musiq}, and ManIQA~\cite{yang2022maniqa}. 
(3) MLLM-based score prediction methods, including CLIP-IQA+~\cite{wang2023exploring}, C2Score~\cite{zhu2024adaptive}, Q-Align~\cite{wu2023q}, and DeQA~\cite{you2025teaching}. 
(4) MLLM-based frameworks with reasoning, such as Q-Insight~\cite{li2025q}, VisualQuality-R1~\cite{wu2025visualquality}, and Zoom-IQA~\cite{liang2026zoom}.

\begin{figure}[t]
\centering
\includegraphics[width=1\linewidth]{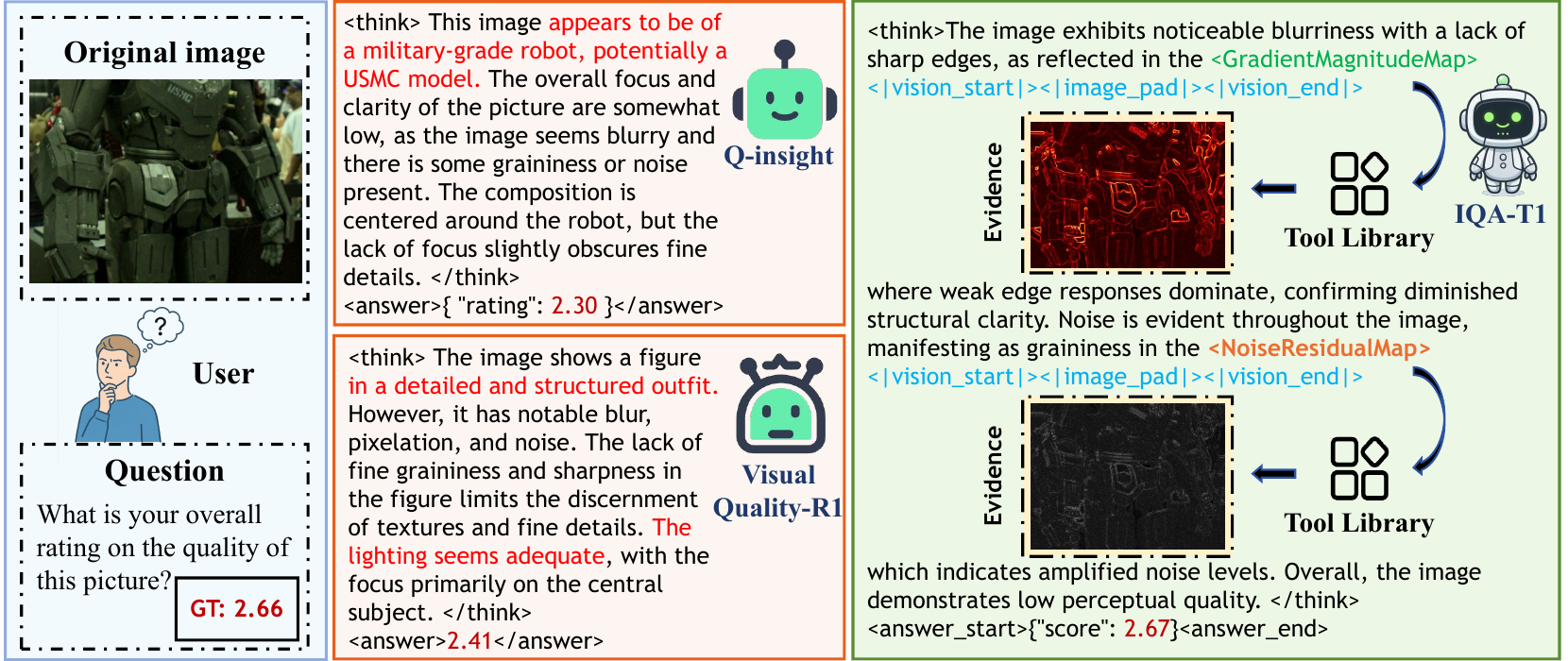}
\caption{Qualitative comparison of \name with competing methods. We highlight: \textcolor{red}{incorrect descriptions}, and the visual evidence reasoning unique to our model.}
\label{fig:case}
\end{figure}

\subsection{Main Results}
\label{sec:main_results}

Table~\ref{tab:plcc_srcc} presents a comprehensive comparison of PLCC and SRCC across all methods and datasets. Our proposed \name achieves the highest average PLCC (0.795) and SRCC (0.784) across the seven datasets.

\noindent\textbf{Comparison with MLLM-based methods without reasoning.}
Methods such as Q-Align~\cite{wu2023q} and DeQA~\cite{you2025teaching} focus exclusively on score regression and achieve strong performance, particularly on in-distribution datasets like KonIQ~\cite{hosu2020koniq}. However, they lack the ability to explain their judgments. \name not only remains highly competitive in score accuracy but also generates interpretable, evidence-grounded reasoning chains, offering a unique combination of precision and transparency.

\noindent\textbf{Comparison with MLLM-based reasoning methods.} Among methods that produce both scores and rationales, \name exhibits strong cross-distribution generalization (Qualitative examples are shown in Fig.~\ref{fig:case}). It demonstrates robust performance on real-world images (e.g., SPAQ~\cite{fang2020perceptual}), establishes new highest results on challenging datasets for typical synthetic and algorithm-induced distortions (CSIQ~\cite{larson2010most} and PIPAL~\cite{jinjin2020pipal}, respectively), and remains highly competitive on AI-generated content (AGIQA~\cite{li2023agiqa}).

 These results provide strong empirical evidence for the central thesis of this work: by augmenting MLLM reasoning with explicit, tool-generated visual evidence, \name effectively reduces the semantic bias inherent in standard MLLM representations.

\subsection{Ablation Studies and Analysis}
\label{sec:ablation}

\begin{figure}[t]
\centering
\includegraphics[width=1\linewidth]{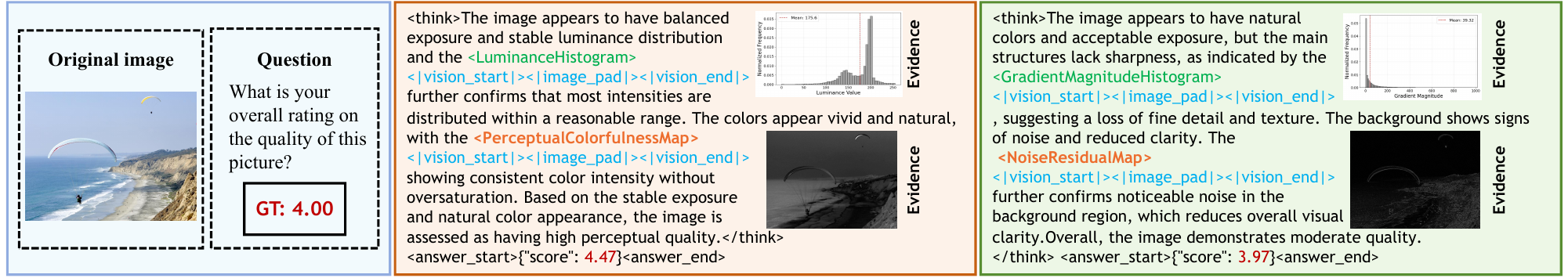}
\caption{Qualitative comparison of \name with competing methods.}
\label{fig:wrong-case}
\end{figure}

\begin{table}[t]
\centering
\Huge
\caption{Ablation studies of individual components measured by PLCC and SRCC, where all models are trained using the KonIQ dataset.}
\label{tab:Ablation}
\renewcommand{\arraystretch}{1.15}
\setlength{\tabcolsep}{4.5pt}
\resizebox{\textwidth}{!}{
\begin{tabular}{c c c c|c c c c c c c c}
\toprule
\textbf{Tool} & \textbf{SFT} & $\mathbf{R_{\text{tool}}}$ & $\mathbf{R_{\text{rep}}}$
& \textbf{KonIQ~\cite{hosu2020koniq}} 
& \textbf{SPAQ~\cite{fang2020perceptual}} 
& \textbf{LiveW~\cite{ghadiyaram2015live}} 
& \textbf{KADID~\cite{lin2019kadid}} 
& \textbf{PIPAL~\cite{jinjin2020pipal}} 
& \textbf{CSIQ~\cite{larson2010most}} 
& \textbf{AGIQA~\cite{li2023agiqa}} 
& \textbf{AVG.} \\
\midrule

$\times$ & $\checkmark$ & $\times$ & $\times$
& 0.843 / 0.817
& 0.848 / 0.836
& 0.789 / 0.742
& 0.621 / 0.603
& 0.412 / 0.418
& 0.675 / 0.640
& 0.747 / 0.701 
& 0.705 / 0.680\\

$\checkmark$ & $\checkmark$ & $\times$ & $\times$
& 0.915 / 0.901
& 0.891 / 0.890
& 0.826 / 0.774
& 0.660 / 0.705
& 0.464 / 0.455
& 0.768 / 0.730
& 0.770 / 0.712 
& 0.756 / 0.738\\

$\checkmark$ & $\checkmark$ & $\checkmark$ & $\times$
& 0.933 / 0.917
& 0.903 / 0.904 
& 0.881 / 0.858
& 0.684 / 0.704 
& 0.479 / 0.481 
& 0.841 / 0.835
& 0.811 / 0.739 
& 0.790 / 0.777\\

$\checkmark$ & $\checkmark$ & $\times$ & $\checkmark$
& 0.935 / 0.919
& 0.902 / 0.902
& 0.874 / 0.840
& \textbf{0.689} / \textbf{0.708}
& \textbf{0.481} / 0.485 
& 0.838 / 0.832
& 0.804 / 0.742 
& 0.789 / 0.775\\

$\checkmark$ & $\checkmark$ & $\checkmark$ & $\checkmark$
& \textbf{0.942 / 0.925}
& \textbf{0.905 / 0.908}
& \textbf{0.888 / 0.860}
& 0.686 / \textbf{0.708}
& 0.480 / \textbf{0.490}
& \textbf{0.850 / 0.840}
& \textbf{0.816 / 0.754}
& \textbf{0.795 / 0.784}\\

\bottomrule
\end{tabular}
}
\end{table}

\begin{table}[t]
\centering
\Huge
\caption{Ablation study of different tool invocation strategies on the KonIQ dataset.}
\label{tab:dynamic_tool}
\renewcommand{\arraystretch}{1}
\setlength{\tabcolsep}{4.5pt}
\resizebox{0.78\textwidth}{!}{
\begin{tabular}{l | c c c c c}
\toprule
\textbf{Method} 
& \textbf{PLCC} 
& \textbf{SRCC} 
& \textbf{Avg. Tools} 
& \textbf{Avg. Time(s)} 
& \textbf{Peak Mem.(GB)} \\
\midrule

Qwen3-VL-4B & 0.785 & 0.744 & 0.00 & \textbf{4.62} & \textbf{8.52} \\
\name (All Tools) & 0.865 & 0.841 & 15.00 & 6.06 & 9.13 \\
\name (Fixed-$K$) & 0.895 & 0.869 & 3.00 & 5.32 & 8.70 \\
\name (Random-$K$) & 0.871 & 0.854 & 3.00 & 5.42 & 8.74 \\
\name (Dynamic) & \textbf{0.942} & \textbf{0.925} & 2.34 & 5.09 & 8.67 \\

\bottomrule
\end{tabular}
}
\end{table}

\begin{table}[t]
\centering
\Huge
\caption{Generalization of \name across different MLLM backbones. Results are reported as PLCC/SRCC.}
\label{tab:backbone_generalization}
\renewcommand{\arraystretch}{1}
\setlength{\tabcolsep}{5pt}
\resizebox{\textwidth}{!}{
\begin{tabular}{l | c c c c c c c}
\toprule
\textbf{Method}
& \textbf{KonIQ~\cite{hosu2020koniq}}
& \textbf{SPAQ~\cite{fang2020perceptual}}
& \textbf{LiveW~\cite{ghadiyaram2015live}}
& \textbf{KADID~\cite{lin2019kadid}}
& \textbf{PIPAL~\cite{jinjin2020pipal}}
& \textbf{CSIQ~\cite{larson2010most}}
& \textbf{AGIQA~\cite{li2023agiqa}} \\
\midrule

\name (Qwen3-VL-2B)
& 0.929 / 0.910
& 0.892 / 0.898
& 0.851 / 0.800
& 0.654 / 0.700
& 0.471 / 0.470
& 0.812 / 0.773
& 0.792 / 0.723 \\

\name (InternVL3.5-4B)
& 0.898 / 0.897
& 0.881 / 0.875
& 0.842 / 0.821
& \textbf{0.701 / 0.719}
& 0.443 / 0.437
& 0.749 / 0.733
& \textbf{0.822 / 0.774} \\

\name (Qwen3-VL-4B)
& \textbf{0.942 / 0.925}
& \textbf{0.905 / 0.908}
& \textbf{0.888 / 0.860}
& 0.686 / 0.708
& \textbf{0.480 / 0.490}
& \textbf{0.850 / 0.840}
& 0.816 / 0.754 \\

\bottomrule
\end{tabular}
}
\end{table}

We provide a set of complementary analyses to better understand the effectiveness and generality of \name. Table~\ref{tab:Ablation} isolates the contribution of individual components, Table~\ref{tab:dynamic_tool} compares different tool invocation strategies and their inference costs, and Table~\ref{tab:backbone_generalization} evaluates the robustness of the proposed framework across different MLLM backbones.

\noindent\textbf{Effectiveness of Visual Evidence.}
We first examine the effect of tool-generated visual evidence by comparing two SFT variants: one where tool calls return only placeholders without actual evidence (Table~\ref{tab:Ablation} Row 1), and another where visual evidence is available during reasoning (Table~\ref{tab:Ablation} Row 2). Since both variants share identical training objectives and reasoning templates, the difference isolates the impact of visual evidence. The results show consistent improvements across all datasets, indicating that structured visual evidence effectively compensates for the low-level information loss in $\mathbf{z}_v$ and mitigates the semantic bias discussed in Sec.~\ref{sec:problem}.

\noindent\textbf{Effectiveness of Dynamic Tool Invocation.}
We compare dynamic tool invocation with all, fixed, and random tool selection strategies in Table~\ref{tab:dynamic_tool}. Compared with the base Qwen3-VL-4B, tool-generated evidence brings clear gains with moderate inference overhead. However, more evidence is not necessarily better, as irrelevant tools may introduce redundant visual tokens and distracting cues. In contrast, dynamic invocation adaptively selects relevant evidence for each image, achieving the best performance with fewer tool calls. The qualitative case study in Fig.~\ref{fig:wrong-case} further shows that tools aligned with dominant degradations provide stronger support for quality judgment, indicating that \name benefits from evidence reasoning rather than simple score fitting.

\noindent\textbf{Effectiveness of Reward Design.}
We next analyze the contribution of the RL-stage reward components. Starting from the full model, we ablate either the tool usage reward $R_{\text{tool}}$ or the repetition penalty $R_{\text{rep}}$ in Table~\ref{tab:Ablation}. Removing either component leads to degraded performance, indicating that both rewards are necessary for stable tool-based reasoning. Specifically, $R_{\text{tool}}$ encourages the model to acquire sufficient perceptual evidence, while $R_{\text{rep}}$ suppresses redundant or repetitive tool calls. Their combination enables the model to balance evidence sufficiency and evidence diversity, leading to more reliable quality prediction.

\noindent\textbf{Generality across Backbones.} To examine whether the proposed framework depends on a specific MLLM architecture, we instantiate \name with different backbones, including Qwen3-VL-2B, InternVL3.5-4B, and Qwen3-VL-4B. As shown in Table~\ref{tab:backbone_generalization}, \name maintains strong performance across different model scales and architectures. Although the absolute performance varies with backbone capacity and pretraining characteristics, the overall trend remains consistent across real-world, synthetic, algorithm-induced, and AI-generated image quality benchmarks. These results suggest that tool-based visual evidence reasoning is a general mechanism for enhancing MLLM-based IQA, rather than a backbone-specific design.

\section{Conclusion}

In this work, we introduced \name, a tool-based visual evidence reasoning framework that grounds image quality assessment in explicit perceptual evidence rather than semantically biased internal representations. By enabling MLLMs to autonomously invoke specialized analysis tools, our approach transforms IQA from purely semantic inference into an evidence-driven reasoning process.
Supported by the proposed \data dataset and a two-stage training strategy combining supervised learning and reinforcement learning, \name achieves the best average performance across multiple IQA benchmarks while providing interpretable reasoning grounded in visual evidence.
We hope this work encourages future research on integrating structured visual evidence into multimodal reasoning systems, advancing more reliable and interpretable perception in multimodal large language models.


\section*{Acknowledgements}
This work was supported in part by the National Natural Science Foundation of China (No. 62472359, No. 62372379), in part by Xi’an’s Key Industrial Chain Core Technology Breakthrough Project: AI Core Technology Breakthrough under Grand 24ZDCYJSGG0003, and in part by the Hong Kong University of Science and Technology under Grant No. WEB26EG02.

\bibliographystyle{splncs04}
\bibliography{ref}
\end{document}